\documentclass{article}
\usepackage{amssymb}
\usepackage{adjustbox}
\usepackage{booktabs}
\usepackage{dirtytalk}
\usepackage{enumitem}
\usepackage{spconf,amsmath,graphicx}
\usepackage[hidelinks]{hyperref}
\usepackage[noabbrev,capitalize,nameinlink]{cleveref}
\usepackage[shrink=25]{microtype}  %
\usepackage{tcolorbox}
\usepackage{xcolor}
\usepackage[T1]{fontenc}
\usepackage{amsmath}

\crefname{section}{\S}{\S\S}
\Crefname{section}{\S}{\S\S}    %
\crefformat{section}{#2\S#1#3}  %
\Crefformat{section}{#2\S#1#3}
\crefrangeformat{section}{\S\S#3#1#4--#5#2#6}
\Crefrangeformat{section}{\S\S#3#1#4--#5#2#6}
\crefformat{section}{#2\S#1#3}  %
\crefrangeformat{section}{\S\S#3#1#4--#5#2#6}

\newcommand{\baseline}[1]{\textsc{#1}}

\newcommand{\slt}{spoken language translation}
\newcommand{\Slt}{Spoken language translation}

\newcommand{\citep}[1]{\cite{#1}}

\definecolor{primary}{HTML}{EA4335}
\definecolor{accent}{HTML}{C5221F}
\definecolor{background}{HTML}{F5F5F5}

\title{Improved Long-Form Spoken Language Translation with Large Language Models}
\name{Arya D. McCarthy$^{\star \dagger}$\footnote{Work performed while the author was an intern at Google Research}%
\qquad Hao Zhang$^{\dagger}$ \qquad Shankar Kumar$^{\dagger}$ \qquad Felix Stahlberg$^{\dagger}$ \qquad Axel H. Ng$^{\dagger}$}
  
\address{$^{\star}$Center for Language and Speech Processing, Johns Hopkins University \\
  $^{\dagger}$Google Research %
}
\begin{document}
\maketitle
\begin{abstract}
A challenge in \slt{} is that plenty of spoken content is long-form, but short units are necessary for obtaining high-quality translations.
To address this mismatch, we fine-tune
a general-purpose, large language model to split long ASR transcripts into segments that can be independently translated so as to maximize the overall translation quality.
We compare to several segmentation strategies and find that our approach improves BLEU score on three languages by an average of 2.7 BLEU overall compared to an automatic punctuation baseline.
Further, we demonstrate the effectiveness of two constrained decoding strategies to improve well-formedness of the model output from above 99\% to 100\%.
\end{abstract}
\section{Introduction}
\label{sec:intro}

The machine translation component of a cascade model for \slt{}~\citep{cascadeslt} typically operates on sentence-like units, with each sentence translated independently of the others. When asked to translate long sentences, the models regularly fail or degenerate \citep{cho-etal-2014-properties,pouget-abadie-etal-2014-overcoming,koehn-knowles-2017-six}.
This differs considerably from the expectations for automatic speech recognition models (e.g. \citep{rnnt}) that can process inputs of unbounded lengths.

This divergence poses a challenge for \slt{} cascades. They must either be able to cope with potentially long, multi-sentence inputs or, alternatively, they must be able to determine cutpoints at which the transcript
can be segmented into compact, independently translatable units.

Being able to directly accommodate long, multi-sentence inputs has obvious allure. Document- or discourse-level context can be preserved \citep{inaguma2021espnet}, and diction and prosody can affect translated units \citep{tiedemann-scherrer-2017-neural}.
However, in practice, the training data available for the translation models used in cascades is dominated by the relatively short sentence pairs that are preferentially recovered by standard sentence alignment approaches \citep{gale-church-1993-program,thompson-koehn-2019-vecalign}. %
Unfortunately, generalization from training on short sentences continues to be an unsolved problem even in otherwise effective translation models \citep{koehn-knowles-2017-six}, and Transformer models struggle with long context windows \citep{DBLP:journals/corr/abs-2004-05150}. %
Rather than addressing the length generalization problem directly, our work side-steps this by segmenting transcripts into units that maximize the performance of a machine translation system that operates optimally only on short segments of input text.

While numerous text segmentation techniques have been proposed to improve \slt{} (\cref{sec:related-work}), 
the problem remains hard and unsolved.
Indeed, Li et al.~\citep{li-etal-2021-sentence} demonstrate that poor
sentence segmentation degrades performance almost twice as
much as transcript level-errors. Moreover, when the first three authors each attempted to manually sentence-segment one of the 14 passages in our development set, we achieved F1 scores of 0.68, 0.79, and 0.85, evincing the ambiguity in the task.

Our approach combines a fine-tuned Text-to-Text Transfer Transformer (T5) model \citep{t5} and windowed input--output pairs. We show comprehensive comparisons between various baselines and decoding strategies.
Experiments in three language pairs indicate that our approach outperforms a baseline that carries out no sentence splitting, a cascade system that predicts punctuation marks before inferring sentence boundaries, and a strong autoregressive neural model.
Overall, we improve the BLEU score on the IWSLT test sets by 2.7 BLEU, 
as a component of a complete \slt{} cascade.\looseness=-1

\begin{figure*}[htb]
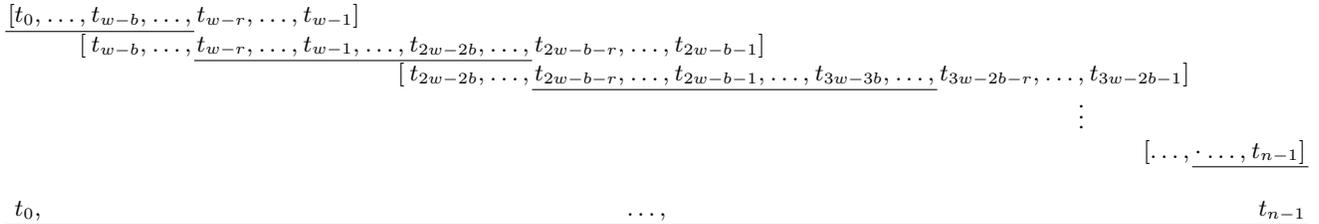

  \centering
\(
\small{  
\arraycolsep=1.0pt\def\arraystretch{1.0}
\begin{array}{rrrrrrrrrrrrrrrrrrrrrr}
  [t_0, & \ldots, & t_{w-b}, & \ldots, & t_{w-r}, & \ldots, & t_{w-1}] & & & & & & & & & & & & &  &\\
  \cline{1-4}
        &        [& t_{w-b}, & \ldots, & t_{w-r}, & \ldots, & t_{w-1}, & \ldots, & t_{2w-2b}, & \ldots, & t_{2w-b-r}, & \ldots, & t_{2w-b-1}] & & & & & & &  &\\
                                      \cline{5-10}
        &         &         &         &         &         &         &       [ & t_{2w-2b}, & \ldots, & t_{2w-b-r}, & \ldots, & t_{2w-b-1}, &\ldots, & t_{3w-3b}, & \ldots, & t_{3w-2b-r}, & \ldots, & t_{3w-2b-1} ]   & \\
                                                                                                     \cline{11-16}
                                      
        &         &         &         &         &         &         &         &          &         &          &         &          &        &         &          &          & \vdots  &             & & &\\                                                      &         &         &         &         &         &         &         &          &         &          &         &          &        &         &          &          &         & [\ldots, & \cdot & \ldots, & t_{n-1}]    \\
                                                                                                     \cline{20-22} \\

                                                                                                     \hphantom{[}t_0, & \multicolumn{20}{c}{\ldots,} & t_{n-1}\\                                                                                                     
\cline{1-22}                                                                                                     
\end{array}
}
\)
\caption{Processing overlapping windows instead of entire transcript passages. $w$ is the window size used in both training and inference. $b$ is the total context window size. $r$ ($\le b$) is the right context window size. The underlines below the windows indicate which local segmentation decisions are taken as global decisions. Portions not underlined (i.e., the context window) are still provided to the segmentation model to inform segmentation of underlined portions. %
}
  \label{fig:windowing}
\end{figure*}

\section{Problem Setup}
The input is a sequence of $n$ word tokens from an automatic speech recognition (ASR) system. The output is a sequence of binary segmentation decisions: $\mathcal{Y} = \{\textsc{split},\textsc{continue}\}$ with $y_t \in \mathcal{Y}$ for ($0 \leq t < n$). T5 encodes this as a copy of the input sequence with delimiters between certain words.

One major challenge of modeling and inference of long-form transcript segmentation is that the input sequences can be very long. For example, a TED talk can contain more than one thousand words ~\citep{li-etal-2021-sentence}.
We take a divide-and-conquer approach that operationalizes two straightforward principles in modeling.
First, words on the left and right are both useful for deciding if a sentence delimiter should be present at the current word position.
Second, distant words are less useful than nearby words. From these two principles, we design a top-level sliding window algorithm to balance the need for bidirectional modeling and efficiency of computation.
We divide the passage into windows at both training and test time, with a small context window on each side to inform decisions at window edges (\cref{fig:windowing}).
With this top-level inference algorithm, the sequence-to-sequence machine learning problem is now reduced to the window-level. The problem is now to predict a sequence of segmentation decisions $\mathbf{y}=y_1,\ldots,y_{w}$ for each text \emph{window} of size at most $w$ tokens: $\mathbf{x}=x_1,\ldots,x_{w}$. %

\section{Modeling Approaches}

A classic approach to discriminative sequence modeling is the conditional random field (CRF) \citep{lafferty-etal-2001-conditional,liu-etal-2005-using-conditional}. This conditional graphical model allows incorporating arbitrary features of the transcript, including linguistic variables and word embeddings. %

\subsection{Bidirectional RNN Model}
\label{sec:birnn_model}
The limitation of the CRF is in the Markov assumption it makes, considering only the immediately previous word's segmentation decision.
Even higher-order CRFs can only consider a fixed-size history within $\mathbf{y}$. 
Instead, we introduce a neural autoregressive segmenter. It is an encoder-decoder neural network with monotonic hard attention to the bidirectionally encoded input at the current word position, admitting the same rich featurization of $\mathbf{x}$ as the CRF; its likelihood is
\[
\arraycolsep=1.0pt\def\arraystretch{1.5}
\begin{array}{rcl}
  p_\theta(\mathbf{y} \mid \mathbf{x}) &=& \prod_{t=1}^w p_\theta(y_{t} \mid \mathbf{y}_{<t}, \mathbf{x})  \\
                                      &:=& \prod_{t=1}^w p_\theta(y_{t} \mid \mathbf{y}_{<t}, \mathbf{BiRNN(x)}_t) \\
\end{array}
\]
where $p_\theta$ is parameterized by a recurrent neural network followed by a linear projection layer and a softmax layer to obtain a locally normalized distribution.
Exact inference here is intractable (unlike a CRF); we approximate it with beam search.

\subsection{Text-to-Text Transfer Transformer (T5)}
More recently, the paradigm of pre-training followed by fine-tuning has achieved great successes across many NLP tasks. The pre-training task is typically  a variant of a language model \cite{gpt,palm} or an autoencoder \cite{t5} where a corrupted version of a sentence is mapped to its uncorrupted counterpart. We can encode segmentation as such a task: reproducing the input with inserted sentence delimiters.
Concretely, we encode $\mathbf{y}$ as $z_0,\ldots,z_{w-1}$ where $z_t = \text{Concat}(d_t,x_i)$ and $d_i \in \{\epsilon, \blacksquare\}$.
For example, we feed \texttt{i am hungry i am sleepy} to the model, and it produces the sentence-delimited string \texttt{i am hungry $\blacksquare$ i am sleepy}.
We use the publicly available T5 model \citep{t5} as the foundation for our text-based segmenter.  %

A deficiency of generation with T5 (which we observed in other text-based models during development) is that the output might not only fail to correctly segment the passage; it might not even contain the same tokens as the passage. We shall say that an output is \emph{well-formed} if it contains the same token sequence as the input, with zero or one sentence delimiters before each token.
While the rich parameterization of such large Transformer models may learn the inherent structure of the output, we also provide two solutions to enforce well-formedness. 
As a third option, we use T5 to rerank the $n$-best hypotheses from the BiRNN, combining well-formedness constraints with efficient modeling.

\subsubsection{Finite State Constraints in Decoding}
\label{sec:fst_constraint}

\begin{figure*}[htb]
    \centering
    \includegraphics[width=0.95\linewidth]{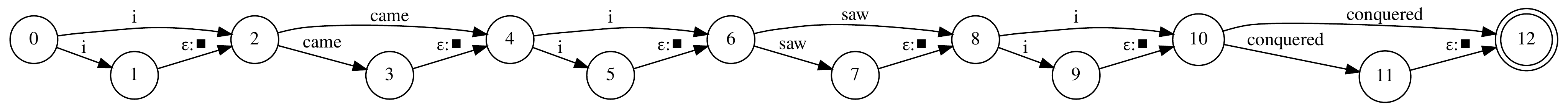}
    \caption{FST representing all possible segmentations for the transcript \protect\say{i came i saw i conquered}.
      }
    \label{fig:sawtooth}
\end{figure*}

A natural strategy to force well-formed outputs is \emph{constrained decoding} (e.g. \cite{zhang-etal-2019-neural}).
In it, we compose the input FSA $\mathbf{x}$ and a special FST $\mathcal{T}$ encoding all possible segmentation decisions, then project the FST to the output tape to obtain a determinized FSA for the output space. The FST $\mathbf{x} \circ \mathcal{T}$ is shown in \cref{fig:sawtooth}.

\subsubsection{Levenshtein Alignment for Post-processing} \label{sec:levenshtein}
\label{sec:levenshtein_alignment}
The generation models' ability to produce arbitrary outputs may be seen as a strength: the model could correct transcription errors and remove disfluencies, if so trained.
Therefore, alternatively we can let the model generate freely without enforcing structural constraints, then enforce well-formedness post-hoc.
Kumar and Byrne \citep{kumar02_icslp} describe a WFST for \textit{Levenshtein alignment} between two strings. We use it to align the generated string with $\mathbf{x}$. We then project segment boundaries across alignment links from the generated string onto $\mathbf{x}$ to determine $\mathbf{y}$. In this way, annotations can be salvaged when T5 does not precisely recreate the input.

\section{Experiments}
Our experiments are carried out on the IWSLT spoken language translation data sets, subjected to the same pre-processing as described in Li et al.~\citep{li-etal-2021-sentence}. We use the 2014 data for dev and 2015 and 2018 for test.
The fourteen reference transcripts in our dev set range from 861 to 1234 words; by contrast, the median length of a sentence in written English is close to 17 words \citep{kucera1970computational}. %
We use the publicly available Speech-to-Text Google API (\url{https://cloud.google.com/speech-to-text}) %
to generate ASR transcripts.
We remove punctuation and lowercase the ASR transcripts and use English-\{German,Spanish,Arabic\} machine translation models trained with the same preprocessing on the source side as~\citep{li-etal-2021-sentence}.

The MT model is a Transformer with a model dimension of 1024, hidden size of 8192, 16 attention heads, 6 encoder layers, and 8 decoder layers. We decode with a beam size of 4. 
In our experiments, the three MT model instances and the ASR model (and thereby its transcripts) are fixed while we vary the sentence segmentation policies. %
We report case-sensitive BLEU as computed with \textsc{sacrebleu}~\citep{post-2018-call}.
We also report the F1 score of segment boundary prediction for diagnostic purposes. %

We compare our proposed solution's performance to several baselines which illustrate natural
first-pass options for segmentation decisions in \slt{}.

\baseline{FixedLength} %
    separates the transcript into disjoint segments with the same number of tokens. While this requires no external segmentation model, the resulting segments are non-sentential \citep{tsiamas2022shas}.
    
    \baseline{Punctuate} %
    is a two-pass segmentation that first infers punctuation with a pQRNN \citep{soboleva2021replacing}, then uses a fixed set of inference rules to distinguish sentence-terminal punctuation marks from sentence-internal ones as in “St. John” and “The end.”

    \baseline{Oracle} %
    uses punctuation from the reference transcripts to segment. The segmentation is projected onto Levenshtein-aligned words in the noisy ASR transcripts (\cref{sec:levenshtein}). (A true oracle would optimize corpus-level BLEU over all $2^n$ segmentations, but this is intractable.)

  \baseline{BiRNN}
  is a shallow BiRNN model trained on the C4 data set \citep{t5} using the same rules in \baseline{Punctuate} to derive sentence boundaries as supervision.
  The model has 1 left-to-right GRU layer, 1 right-to-left GRU layer, and 1 GRU layer in the decoder. It uses embeddings of character $n$-gram projections \citep{zhang-etal-2019-neural}. 
  
  \baseline {BiRNN f.t.}
  fine-tunes on the IWSLT training set, validated on the dev set, after training on C4.
  
\begin{figure}
    \centering
    \includegraphics[width=0.8\linewidth]{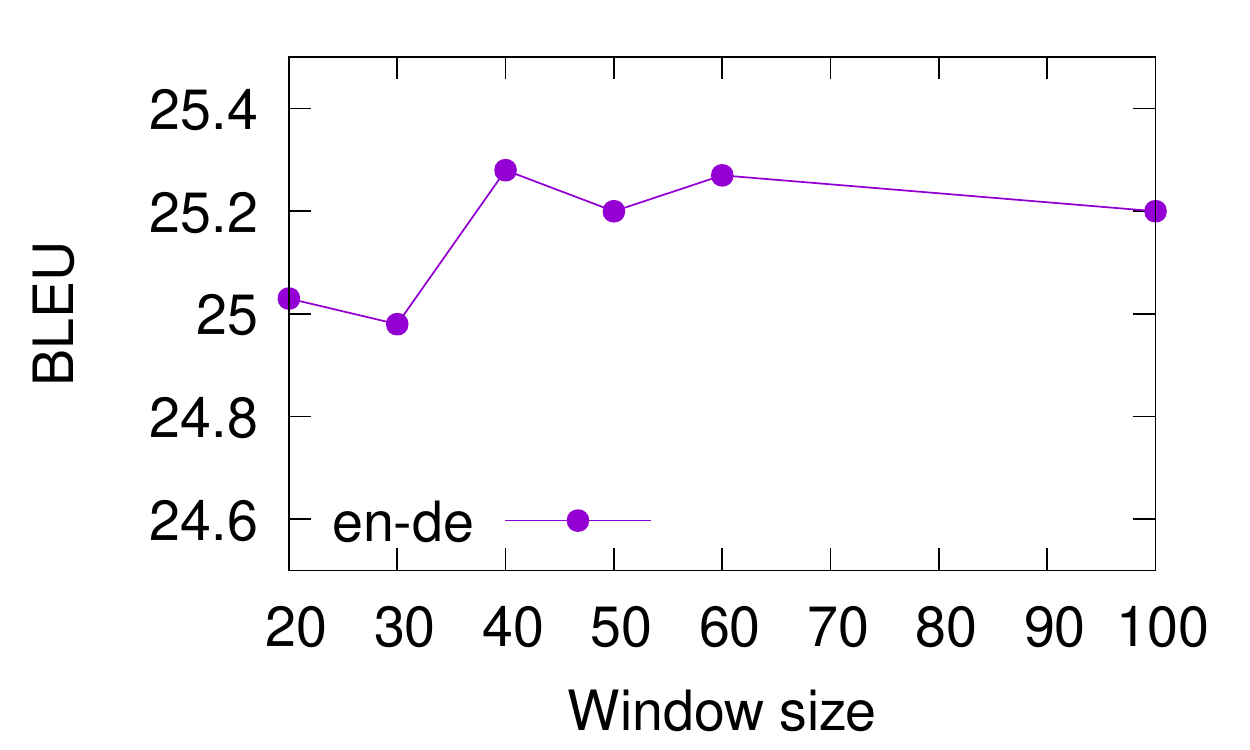}
    \caption{T5 segmentation BLEU as window size increases.}
    \label{fig:window-bleu}
\end{figure}

  \baseline {T5-base} and \baseline {T5-11B}
  fine-tune the base T5 model and 11B T5 model (xxl) respectively~\citep{t5} on the IWSLT train and dev sets.
\Cref{fig:window-bleu} shows that T5 is insensitive to the length of input windows defined in \cref{fig:windowing}, except for very short ones. Smaller windows are faster for training and inference, so we use 40-word windows with left and right context of 5.

\section{Results}
\begin{table*}[t]
    \centering
    \begin{adjustbox}{max width=0.75\linewidth}
    \begin{tabular}{@{} l r r r r r r r r r r @{}}
    \toprule
     &F1 & \multicolumn{3}{c}{\textsc{en-de}} & \multicolumn{3}{c}{\textsc{en-es}} & \multicolumn{3}{c}{\textsc{en-ar}} \\
     \cmidrule(lr){3-5} \cmidrule(lr){6-8} \cmidrule(lr){9-11}
     Policy  & TED 2014 &  2014 & 2015 & 2018 &  2014 & 2015 & 2018 &  2014 & 2015 & 2018 \\
     \midrule
     \baseline{Oracle}        & 1.000  & 26.66  & 30.24  & 25.21  & 40.38  & 41.72  & 41.84 & 15.66 & 18.18 & 17.59\\
     \baseline{FixedLength}   & 0.041  & 20.82  & 23.45  & 19.66  & 32.76  & 34.03  & 34.01 & 12.64 & 14.79 & 13.92 \\
     \baseline{Punctuate}     & n/a    & 22.80  & 26.30  & 21.60  & 35.70  & 36.90  & 36.70 & 13.70 & 15.80 & 15.40\\     
    \addlinespace
    \baseline{BiRNN}          & 0.669  & 24.43  & 27.72  & 22.42  & 36.83  & 38.37  & 38.04 & 14.38 & 16.56 & 15.98 \\
    \baseline{BiRNN f.t.}     & 0.697  & 24.55  & 28.10  & 23.14  & 37.31  & 39.08  & 38.64 & 14.41 & 16.77 & 16.19 \\
    \addlinespace
    \baseline{T5-base}        & 0.788  & 25.28 & 29.14   & 24.05    & 38.75  & 40.23  & 39.96   & 14.94 & 17.32  & \textbf{16.57}\\
    \baseline{T5-11B}         & \textbf{0.821}  & \textbf{25.63} & \textbf{29.63}   & \textbf{24.27}    & \textbf{39.16}  & \textbf{40.64}  & \textbf{40.05}   & \textbf{15.31} & \textbf{17.60}  & 16.48 \\

    \bottomrule
    \end{tabular}
    \end{adjustbox}
  \caption{Segmentation F1 scores on dev set and BLEU scores on dev and test sets, translating into German, Spanish, and Arabic.
  }
    \label{tab:results}
    
\end{table*}

\Cref{tab:results} compares different segmentation policies when translating the IWSLT test sets.
Fine-tuning is effective for both the shallow BiRNN model and the large T5 model. But for BiRNN, the gain is only around 0.5 BLEU. For T5, fine-tuning turns the general purpose text-to-text model into a highly accurate segmentation model, an average of 1.0 BLEU from the oracle on the test sets. 
(Moreover, we outperform \cite{li-etal-2021-sentence} who experiment on the same \textsc{en-de} data, using data augmentation to achieve BLEU of 27.0 and 22.0 for 2015 and 2018 respectively.) On inspection, most of the differences in outputs can be attributed to T5's punctuation decisions matching the reference's better than the BiRNN model. For instance, T5 accurately captures longer-distance dependencies like complete quotes that the BiRNN segmentation prematurely terminates.

T5 models are several magnitudes larger than the BiRNN models. Moreover, within the T5 family, \baseline{T5-11B} is substantially more accurate than \baseline{T5-base} in terms of both segmentation F1 and BLEU scores for all three language pairs. The strong correlation between accuracy and model size is clear.

\subsection{Constraining and Search Strategies}
The motivation for post-hoc Levenshetein alignment or on-the-fly FST-constrained decoding is that T5's decoder output can deviate from the input, misaligning windows and harming downstream translation. On the dev set, T5 base unconstrained is 99.4\% (643/647) well-formed while T5 11B unconstrained is 99.8\% (646/647) well-formed.
While this shows that the T5 models are highly reliable, the fact that even the 11B model has no well-formedness guarantee echos the call of Sproat \citep{10.1162/coli_a_00439} for combining symbolic methods with neural networks.
In practice, either of the two constraint strategies remains necessary for 100\% well-formed output.

\Cref{tab:search_ablation} offers results of the two constraint strategies combined with four search procedures: greedy search, beam search with sizes 4 (default) and 100, and exact search~\citep{stahlberg-byrne-2019-nmt}. The results indicate that T5 learns the structural constraint accurately and assigns most of the probability mass to a few hypotheses. There is no obvious benefit to more exhaustive search.
\begin{table}
  \center{
  \begin{adjustbox}{max width=0.9\linewidth}
  \begin{tabular}{@{} ll r r @{}}
    \toprule
     \baseline{constraint} &\baseline{search} & \textsc{en-de bleu} & \textsc{F1}\\
    \midrule
    \baseline{FST} (\cref{sec:fst_constraint}) &\baseline{greedy} & 25.31& 0.786 \\
    {} &\baseline{beam=4} & 25.28&  0.788\\
    {} &\baseline{beam=100} & 25.29 & 0.788\\
    {} &\baseline{exact} & 25.28 &  0.788\\  
    \addlinespace
    \baseline{Levenshtein} (\cref{sec:levenshtein_alignment}) &\baseline{greedy} & 25.31&  0.786\\
    {} &\baseline{beam=4} & 25.28&  0.788\\
    {} &\baseline{beam=100} & 25.31& 0.788 \\
    {} &\baseline{exact} &  25.28&  0.788\\
    \bottomrule
  \end{tabular}
  \end{adjustbox}
  }
  \caption{Both constraints offer similar performance for each search strategy.}
  \label{tab:search_ablation}  
\end{table}

\subsection{BiRNN Beam Reranking with T5}
In \cref{tab:t5_reranking}, we also show that for each beam width, T5 reranks the $n$-best list of BiRNN effectively, outperforming it to a greater extent as the beam size is increased. This indicates that T5 is superior to the BiRNN model in modeling increasingly large output spaces while the BiRNN assigns most probability mass to a narrow band of hypotheses. Reranking the BiRNN outputs fails to surpass the T5 generation approach, even at beam size 100.

\begin{table}[htb]
  \center{
  \begin{adjustbox}{max width=0.67\linewidth}
  \begin{tabular}{@{} l r r @{}}
  \toprule
      \textsc{beam} & \textsc{en-de bleu} & \textsc{+ reranking} \\
  \midrule
      10 & 24.43 & 24.91 \\
      50 & 24.43 & 25.12 \\
      100& 24.43 & \textbf{25.24} \\
  \bottomrule
  \end{tabular}
  \end{adjustbox}
  }
  \caption{Reranking the BiRNN beam with T5 enables larger beam sizes to yield better translations.}
  \label{tab:t5_reranking}  
\end{table}

\section{Related Work} \label{sec:related-work}

Segmenting long texts into units suitable for translation has been a recurring topic in MT research \citep{li-etal-2021-sentence,8713737,pouget-abadie-etal-2014-overcoming,doi-sumita-2003-input,Goh2011SplittingLI}.
To bridge the gap between ASR and MT, \citep{li-etal-2021-sentence} address long-form spoken language translation. Claiming that segmentation is the bottleneck, they adapt their MT model to work \emph{with} automatic segmentations, however inaccurate they may be.

We are training our models to minimize the loss of source sentence segmentation. The ultimate objective is improving the downstream translation quality. It is interesting to explore reinforcement learning for segmentation \citep{srinivasan-dyer-2021-better}, but the state space is vast for the long-form segmentation problem compared to prior work on RL-based segmentation.

Finally, one may consider additional sources of data or training examples to improve modeling.
Using prosodic features when they are available is viable \citep{tsiamas2022shas}; however, we show that T5 closes most of the accuracy gap without these. As a contrasting approach, \citep{kumar02_icslp} focus on segmenting an ASR \textit{lattice}, rather than the decoded transcript. Finally, data augmentation \citep{li-etal-2021-sentence,9053406} can complement our approach.

\vspace{0.8em}

\section{Conclusion}

We have addressed a key challenge in \slt{}: the segmentation of transcripts into machine-translatable units.
We introduce a large language model--based approach for sentence segmentation, relying on supervision from only a few thousand long-form transcripts which are readily available.
We propose a sliding window algorithm to handle long-form ASR transcripts efficiently without losing accuracy.
Our model consistently outperforms a pQRNN-based punctuation model by 2.7 BLEU and a strong fine-tuned BiRNN model by 1.3 BLEU on average across three language pairs. %
The best-performing model is within 1 BLEU point of the oracle.

\section{Acknowledgments}
We thank Colin Cherry, Dirk Padfield, and Basak Oztas for early collaborations on the project.  %
A.D.M. is supported by an Amazon Fellowship and a Frederick Jelinek Fellowship.

\bibliographystyle{IEEEbib}
\bibliography{refs}

\end{document}